\documentclass{article}

\usepackage{microtype}
\usepackage{graphicx}
\usepackage{subfigure}
\usepackage{tabularx}
\usepackage{booktabs}
\usepackage{siunitx}

\usepackage{hyperref}

\usepackage[accepted]{icml2023}

\usepackage{amsmath}
\usepackage{amssymb}
\usepackage{mathtools}
\usepackage{amsthm}

\DeclareMathOperator*{\argmin}{arg\,min}

\usepackage[capitalize,noabbrev]{cleveref}

\theoremstyle{plain}

\theoremstyle{definition}

\theoremstyle{remark}

\usepackage[textsize=tiny]{todonotes}

\icmltitlerunning{Submission and Formatting Instructions for ICML 2023}

\begin{document}

\twocolumn[

\icmltitle{Reimplementation of Learning to Reweight Examples for Robust Deep Learning}

\icmlsetsymbol{equal}{*}

\begin{icmlauthorlist}
\icmlauthor{Parth Patil}{equal}
\icmlauthor{Ben Boardley}{equal}
\icmlauthor{Jack Gardner}{equal}
\icmlauthor{Emily Loiselle}{equal}
\icmlauthor{Deerajkumar Parthipan}{equal}
\end{icmlauthorlist}


\icmlcorrespondingauthor{Jack Gardner}{gardne97@purdue.edu}
\icmlcorrespondingauthor{Ben Boardley}{bboardle@purdue.edu}
\icmlcorrespondingauthor{Emily Loiselle}{ealoisel@purdue.edu}
\icmlcorrespondingauthor{Parth Patil}{patil185@purdue.edu}

\icmlkeywords{Machine Learningnn, ICML}

\vskip 0.3in
]


\begin{abstract}
Deep neural networks (DNNs) have been used to create models for many complex analysis problems like image recognition and medical diagnosis. DNNs are a popular tool within machine learning due to their ability to model complex patterns and distributions. However, the performance of these networks is highly dependent on the quality of the data used to train the models. Two characteristics of these sets, noisy labels and training set biases, are known to frequently cause poor generalization performance as a result of overfitting to the training set. This paper aims to solve this problem using the approach proposed by \citet{pmlr-v80-ren18a} using meta-training and online weight approximation. We will first implement a toy-problem to crudely verify the claims made by the authors of \citet{pmlr-v80-ren18a} and then venture into using the approach to solve a real world problem of Skin-cancer detection using imbalanced image dataset.
    
\end{abstract}

\section{Problem Statement}

Label noise in the training of these models is caused mainly by the lack of high-quality labels which can only be produced by extensive human inspection. This misclassification of data can be done either through attacks or unintentional classification. Deep neural networks can be prone to adversarial attacks intended to make the model respond inaccurately, typically in the form of dataset poisoning which is done by inputting carefully crafted data to make the model behave in unintended ways. Additionally, Frenay’s research summarizes the main sources for unintentional noise: insufficient information, subjective classification, and general mistakes ~\cite{Frenay2014}. As for training set biases, they are often caused by a disproportionate representation of certain classes in the labeled dataset which lead to the model weighting these classes differently. This can be formulated as the training set distribution $P(x, y)$ not aligning with the real world or evaluation distribution $P(x^k, y^k)$ ~\cite{pmlr-v80-ren18a}. Training set biases can also take place when the nature of the data is imbalanced, an example of this is in autonomous driving, where, although very important in classification, emergency vehicles and animals are not seen as frequently in the dataset, and thus the model may not be well fine-tuned to these classes. \\

Current solutions to these issues involve small and costly adjustments of hyperparameters to increase or decrease the importance of certain types of labels. Reweighting of training examples has its roots in classical statistical algorithms used to downplay the loss generated by outliers. This methodology of reweighting is also useful in various areas of machine learning like LSTMs and curriculum learning. These methods often have contradictory solutions because they are loss based. These loss based methods optimize hyperparameters by solving the equation below
 
\begin{align}
\lambda^{\ast} &= \argmin_{\lambda} f(\lambda, D_{\text{train}}, D_{\text{test}})
\end{align}

where the function f() measures the loss of a model generated by a specific algorithm with hyperparameters $\lambda$ on training data $\text{D}_{train}$ and evaluated on validation data $\text{D}_{test}$.\\

This hyper-parameter tuning is not only expensive but it can have contradictions when correcting for noise versus training set biases. When trying to correct for label noise many of these algorithms put a larger weight on examples that have smaller training losses as those are typically the cleaner images. However, when accounting for imbalance in datasets, common solutions will aim to give more weight to examples with larger training loss as those are commonly the underrepresented class, which can be seen in many boosting algorithms (e.g. AdaBoost). Due to the contradiction of the solutions, there are large challenges posed by datasets that are both noisy and imbalanced. \\ 

The effects of noise on an imbalanced data set is explored in an experiment in which Medicare Part B fraud detection data is injected with label noise ~\cite{9643276}. This experiment showed that amongst different machine learning algorithms (Linear Regression, Multi-layer Perception, Random Forest and XGBoost) all of them saw significant decreases in performance with injected label noise, including a 20\% drop in Area Under Precision Recall Curve (AUPRC) by the linear regression and an 8\% drop in both XGBoost and Random Forest. The paper being reviewed aims to help provide an efficient solution for datasets of this nature.\\

The proposed solution will aim to provide a more robust approach to solving the problems faced when using lower-quality datasets under the assumption that there is a small high quality validation set that is representative of the task. This validation set is used to guide the example weighting through each training iteration by finding the optimal example weights that minimize the validation loss ~\cite{pmlr-v80-ren18a}. This solution helps solve a research gap as many current solutions to this issue require extensive learned hyper-parameters (1) which can be expensive during training as fine-tuning hyper-parameters is an offline procedure. The proposed solution will help create an online training approach without any additional hyper-parameters.

\section{Method}
The proposed solution aims to provide a more robust approach to solving the problems faced when using biased or noisy labeled dataset, under the assumption that there is a small high quality validation set that is representative of the task. This validation set is used to guide the example weighting through each training iteration by finding the optimal example weights that lead to model parameters that minimize the validation loss ~\cite{pmlr-v80-ren18a}. \\

More formally, the model fits a meta-learning objective that can be implemented along with traditional supervised learning. Assume that there are both a training set $\{(x_i, y_i, 1 <= i <= N\}$ and a validation set $\{(x^v_i, y^v_i, 1 <= i <= M\}$ in which $M << N$. ~\cite{pmlr-v80-ren18a}. The objective of this method is to minimize the weighted expected loss of the model in which $\theta^*(w)$ is the optimal model parameters given the weight vector $\hat{w}$ and a loss function $J(\theta)$.

\begin{equation}\theta^*(w) = \arg\min_{\theta} \sum_{i=1}^{N} w_i J_i(\theta),
\end{equation}

The weighted vector $\hat{w}$ is another optimized parameter. The objective is to find optimal example weights that will lead to the $\theta$ that minimizes the loss on the validation set. In which $J^v_i$ is the loss with respect to the validation set and  $\theta^*(w)$ is with respect to equation 1.

\begin{equation}w^* = \argmin_{w, w \geq 0} \frac{1}{M} \sum_{i=1}^{M} J^v_i(\theta^*(w))).\end{equation}

However, as currently formulated this meta learning objective can be interpreted as nested optimization loops as shown in equation 3, which can be quite expensive as it requires iterations over the dataset for every single loop.

\begin{equation}w^* = \argmin_{w, w \geq 0} \frac{1}{M} \sum_{i=1}^{M} J^v_i(\arg\min_{\theta} \sum_{i=1}^{N} w_i J_i(\theta)).\end{equation}

In order to solve this inefficiency, the paper adopts an online approach to calculating the weights using gradient descent. Following a similar approach to traditional Stochastic Gradient Descent (SGD), this training method involves a descent step for both $\theta$ and the weights, which will be represented by $\epsilon$. Allow $J_{e,n}(\theta) = \epsilon_n * J_n(\theta)$ and allow $B$ to be the current training batch at timestep $t$. Thus allowing us to manipulate traditional SGD to the below equation.
\begin{equation} 
    \theta_{t+1} = \theta_t - \alpha \cdot \frac{1}{ B } \sum^B_{n = 1}\nabla_\theta J_{\epsilon, n}(\theta)
\end{equation}

The optimal weight for example $n$ at time step $t$ can then be estimated by a gradient descent step on $\epsilon_n$ at the same time step with respect the to loss on the validation set of size $M$.
\begin{equation}
    x_{n,t} = -\eta *  \frac{\partial}{\partial \epsilon_{n,t}} \frac{1}{ M } \sum^M_{i = 1} J^v_n(\theta_{t+1}(\epsilon))
\end{equation}

However negative weights can create instability during optimization so $w$ is estimated by.
\begin{equation}
    \widetilde{w}_{n,t} = max(x_{n,t},0)
\end{equation}

Due to the fact that example weights need to add to 1, the weight can be normalized to provide this functionality. In which  $\delta$ is the Dirac delta function to prevent an undefined estimate.

\begin{equation}
    w_{n,t} = \frac{\widetilde{w}_{n,t}}{\lvert\lvert \widetilde{w}_{t} \rvert\rvert_1 + \delta(\lvert\lvert \widetilde{w}_{t} \rvert\rvert_1)}
\end{equation}

Using the gradient descent step on $J^v(\theta)$ with respect to the weights it is possible to reweight examples in a single iteration allowing us to convert the previous meta-learning objective into an online learning procedure. Helping create a robust training implementation without the need to tune hyperparameters offline.

\subsection{Related Work}
This solution helps solve a research gap as many current meta-learning solutions to this issue, such as MAML ~\cite{DBLP:journals/corr/FinnAL17} and few-shot learning ~\cite{pmlr-v80-ren18a}, require learned hyper-parameters which can be expensive during training and thus must be trained offline. The proposed solution will help create an online training approach without any additional hyper-parameters. \\

Another issue among current algorithms that are used to address these distorted datasets is the contradicting approaches for problems with training sets that are noisy and sets that have imbalanced classes. To solve noisy label training sets, classical algorithms tend to favor examples with less training loss because they are more likely to be clean, but solutions to fix class imbalance in datasets, such as hard negative mining ~\cite{MalisiewiczTomasz2011ERfO}, give more weight to examples with higher training loss because they are usually the underrepresented class. By requiring a small unbiased validation set to guide training, so as not to rely on loss calculated completely with the biased training set to set weights, this method allows the proposed solution to resist overfitting noise better compared to previous literature.\\

\subsection{Learning to reweight examples in a MLP (Multi-Layer Perceptron Network)}

This example gives an algorithm to compute $w_{n,t}$ within a MLP network, for which we calculate the gradients of the validation loss w.r.t the local perturbation $(\epsilon)$. The MLP is defined for parameters for each layer, $\theta = (\theta_l)^L_{l=1}$, where L is the number of layers. For each layer, $z^l$ is the pre-activation value computed as a weighted sum of inputs, then a non-linear activation function $\sigma$ is applied to obtain the post-activation $\Tilde{z}_l$ 
\[z_l = \theta^T \Tilde{z}_{l-1}\]
\[\Tilde{z}_l = \sigma(z_l)\]
Now, the algorithm computes the gradients of the validation loss with respect to the perturbation $(\epsilon)$. This calculation involves how adjusting the weights of examples can optimise the training process. These gradients of the validation loss are approximated by local dot products of the gradients of loss w.r.t. $z_l$ $(g_l)$ and the gradients w.r.t. $\theta_l$ $(\Tilde{z}_{l-1}g^T_l)$. 
\[\frac{\partial}{\partial\epsilon_{n,t}}\mathbb{E}[J^v(\theta_{t+1}(\epsilon)|_{\epsilon_{i,t} = 0}]\]
\[ = - \frac{1}{m} \sum \sum (\Tilde{z}^{v_{j,l-1}}{}^T\Tilde{z}_{i,l-1})(g^{v_{j,l}}{}^Tg_{i,l})\]

This equation shows that the gradient is composed on two terms, the difference in pre-activation values between validation and training, and the difference in gradient directions $(g_l)$ for validation and training. This helps us to conclude that, if a pair of training and validation examples are similar and provide similar gradient directions, then the training example is considered to be beneficial and is up-weighted and conversely, if they provide opposite gradient directions, it is down weighted.

\subsection{Implementation using Automatic Differentiation}

The unnormalized weights of examples can be determined by summing up the correlations between the gradients. It leverages automatic differentiation to calculate the gradient of validation loss w.r.t. the example weights in the current batch. This method unrolls the gradient graph of the training batch and employs backward on backward differentiation to obtain second order gradients. 
The process followed by the learning to reweight examples usingautomatic differentiation algorithm is as follows:
\begin{itemize}
    \item Perform a forward pass
    \item Calculate training loss
    \item Perform backward pass and compute gradients w.r.t. model parameters
    \item Calculate validation loss and gradients using automatic differentiation
    \item Update the example weights based in these gradients
    \item Update the model parameters using gradients from the training loss.
\end{itemize}
Because the automatic differentiation method calls for extra computational steps such as two full forward and backward passes on both the validation and training sets to calculate losses and gradients. Also the back on backward pass adds extra complexity. Due to this the total training period is about 3 times longer than with ordinary training, since backward passes also takes the same time as a forward pass.  

\subsection{Proof of Convergence of the Reweighted Training}

    The paper put forth two theorems to prove the convergence properties of the proposed method. The discussion on proofs can be found in the Appendix section of the paper.
    \begin{enumerate}
        \item Theorem 1: The proposed algorithm converges.
        \item Theorem 2: The rate of convergence within $ \epsilon $ is less than $ O(1/\epsilon^2) $
    \end{enumerate}

\section{Experiment}

\subsection{Toy Problem: Implementation Summary}
Sections 3.1-3.4 aim to reimplement the reweighted model using a toy problem. The toy example we use to display the robust effects of this algorithm is an image classification task using the CIFAR-10 dataset. We use a simple convolutional neural network (CNN) model with two different training schemes: the reweighting algorithm and the more traditional stochastic gradient descent (SGD) algorithm. Upon analysis, our replication of the proposed model was able to accurately reproduce the results and generate a more robust model than the traditional model.

\subsection{Toy Problem: Dataset}
The CIFAR-10 dataset includes ten classes, each with 6,000 images. This dataset is commonly used as a benchmark because of its low-resolution images which pose a challenge for models to learn features. Also its diverse set of classes make it suitable for testing the generalization capabilities of models. In our analysis, the distribution of these images were manipulated to simulate biased training sets. 

\subsection{Toy Problem: Implementation Details}
\label {sec:ToyImplementation}
The architecture of our simple CNN model consists of three convolutional layers followed by a max pooling layer to downsample the number of datapoints in the feature map to reduce computational complexity. This is then followed by two fully connected layers. \\

In our implementation we compare the performance of traditional SGD with that of the paper, learning to reweight model parameters. In this experiment we utilize an imbalanced CIFAR10 dataset. In the traditional SGD approach we use 60 epochs with a learning rate of 0.001 and momentum of 0.9. Similarly, the model trained with reweighting uses the same SGD learning rate and momentum. However, in each epoch the algorithm different from the traditional SGD, before we take descent steps on the model parameters we utilize meta-learning to optimize the weights of each sample's loss to minimize the error on the representative validation set. In order to implement meta-learning we leveraged the higher library which allows us to create a meta model that can be used to optimize the weight vector, without effecting the true model's parameters. After optimizing the weights, we then took a gradient descent step on the true model, using the weighted loss. Our code for the implementation can be found in the citations here ~\cite{team19}.

\subsection{Toy Problem: Results}
\begin{figure}[h]
    \centering
    \includegraphics[scale=0.5]{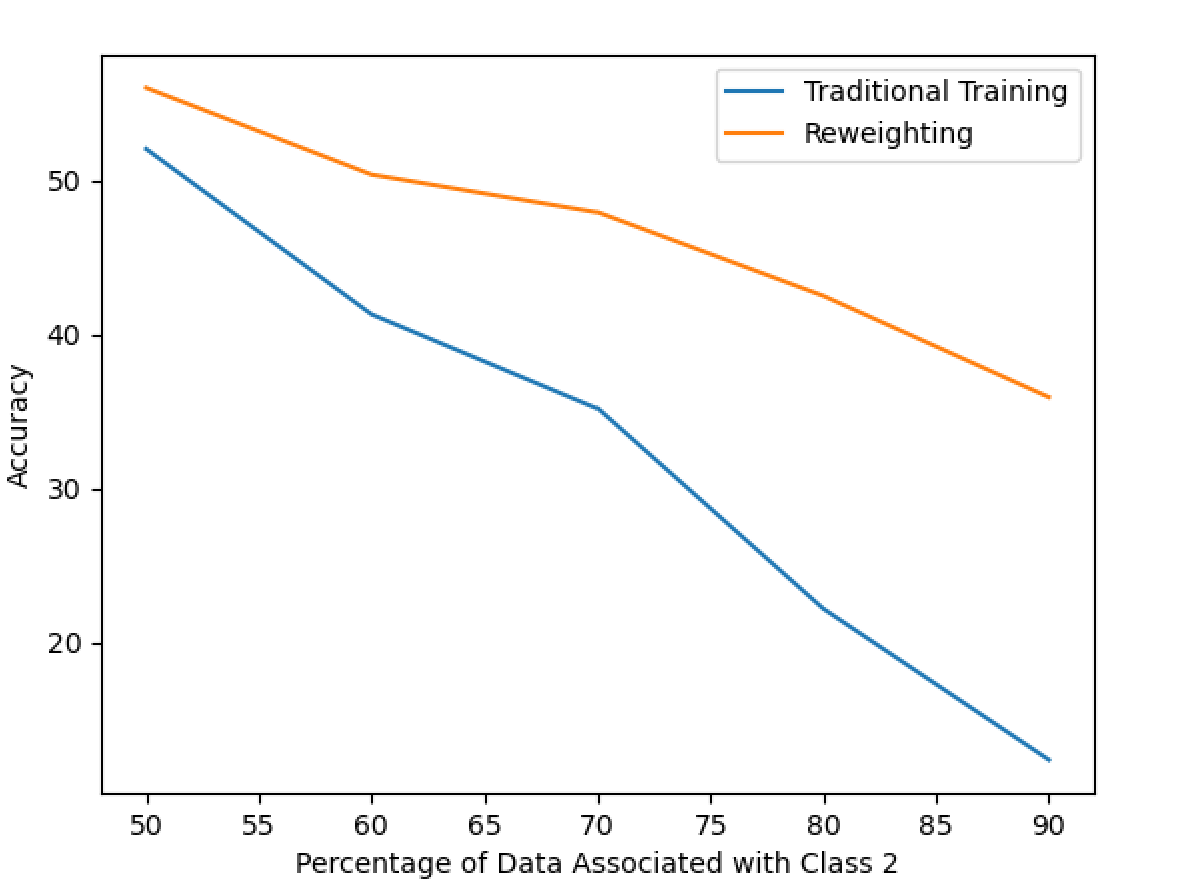}
    \caption{Accuracy of Model Trained with SGD vs Model Trained with Reweighting Algorithm}
    \label{fig:accuracy_fig}
\end{figure}

We analyzed the effects of the reweighting algorithm by simulating different amounts of training set bias for a given class. In Figure~\ref{fig:accuracy_fig}, the reweighting trained model's accuracy is compared to the SGD trained model as the percentage of the training dataset for a single class is increased. For example, at the 60\% mark, 60\% of the training data is associated with a single class while the remaining 40\% is split evenly among the other nine classes. The model that used the reweighting training algorithm was able to out perform the control SGD trained model's accuracy in each test between 50-90\%. It can also be seen that as the amount of class bias increases, the difference in accuracy increases even further.

\subsection{Toy Problem: Conclusion}
The implementation demonstrates the ability of the reweighted method to produce better results on improperly scaled datasets compared to traditional methods. While the accuracy attained by the model in this paper was much lower than the one proposed by \citet{pmlr-v80-ren18a}, this discrepancy can be explained by the fact that our resources only allowed us to train and test on the CIFAR-10 dataset rather than larger equivalent datasets like CIFAR-1000. Additionally, the model in \citet{pmlr-v80-ren18a} was only trained and tested on image recognition tasks with 2 classes rather than 10. Going forward, our next step would be to demonstrate how our model can perform better than traditional models on datasets with increased noise.

\subsection{Real-world Problem Description}
Sections 3.6-3.10 aims to reimplement the reweighted model using a real-world problem. Skin cancer can be deadly if not detected and treated early and accurately. Thus, an accurately trained model that automates diagnosis of suspicious lesions would be able to help patients seek medical intervention sooner, decreasing the risks for patients earlier on. Additionally, automatic diagnosis of skin lesions can help with consistency, healthcare resource allocation, and ease of access to quick diagnoses. \\

Oftentimes, human diagnosis can be subjective and prone to errors, leading to both overdiagnosis and underdiagnosis. Using automated diagnosis models will allow for more consistent and objective assessment. This automated system can then also be easily deployed widely, allowing patients to quickly obtain initial diagnoses without needing appointments. This in turn would be able to help hospitals optimize resource usage, as the need for extensive manual review of screenings and tests is decreased. \\

One issue faced oftentimes when creating an automated diagnosis model is class imbalance in the datasets available. This imbalance is caused because several types of lesions are much more rare compared to other types. In many of the modern algorithms used to train models, this imbalance may cause much higher inaccuracy for these minority classes. Additionally, these datasets can be prone to inaccurate labelling due to human error which can further hurt the accuracy of the model. \\

In order to deal with this class imbalance, we will use the training algorithm described by \citet{pmlr-v80-ren18a} to dynamically adjust the loss function during training.

\subsection{Real-world Dataset}
\begin{figure}[htbp]
    \centering
    \includegraphics[width=0.5\textwidth]{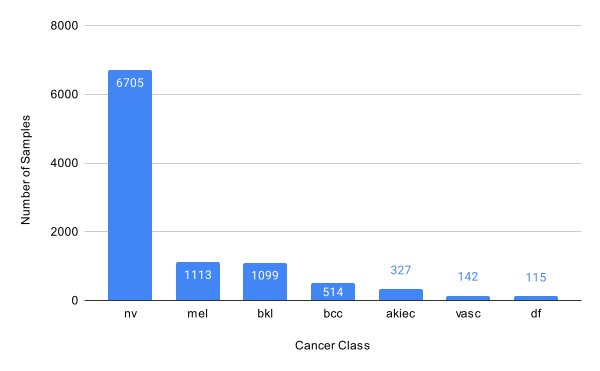}
    \caption{Distribution of training dataset}
    \label{fig:dist}
\end{figure}

The dataset employed for this task is the HAM10000 ("Human against Machine with 10000 training images") dataset. This dataset consists of 10015 dermatoscopic images that contain a characteristic set of 7 different diagnostic classes, collected from multiple populations. These classes include "Actinic keratoses and intraepithelial carcinoma / Bowen's disease (akiec), basal cell carcinoma (bcc), benign keratosis-like lesions (solar lentigines / seborrheic keratoses and lichen-planus like keratoses, bkl), dermatofibroma (df), melanoma (mel), melanocytic nevi (nv) and vascular lesions (angiomas, angiokeratomas, pyogenic granulomas and hemorrhage, vasc)" ~\cite{Ham10000_2018}. This dataset was chosen as our algorithm for reweighting examples should be well equipped to handle a dataset that has the imbalanced nature seen in this dataset. The training samples per class can be seen in Figure ~\ref{fig:dist}.

\subsection{Real-world: Intended Outcome}
Through this experimentation we would like to prove that learning to reweight examples can provide results beyond that of traditional Stochastic Gradient Descent (SGD) and other reweighting algorithms (e.g. boosting algorithms). In addition we hope that our re-implemented model can be competitive with the top 10 performances of task 3, lesion diagnosis, in the 2018 ISIC Competition, as the testing ground truth has been released. The balanced multi class accuracy in the top 10 range from 78\% to 88\% on the leader-board ~\cite{Competition}.

\subsection{Real-world Problem: Experimental Setup}
In our implementation of models built to solve this classification, we use three schemes for training: the reweighting algorithm described in ~\cite{pmlr-v80-ren18a}, a conventional stochastic gradient descent (SGD) algorithm without any weighting, and an algorithm with a static weighted sampler implemented using Pytorch's weighted RandomSampler. For all of these schemes, our models were built by fine-tuning a pretrained Resnet50 model. The un-weighted and weighted sampler schemes were trained with 20 epochs while the model trained with the reweighting algorithm underwent 8 epochs. All 3 models had a learning rate of 0.001 and momentum of 0.9 and used stochastic gradient descent to optimize their parameters. In the weighted sampler model, the imported algorithm ensures that all data points are used during training but adjusts the probability of selecting each class in each of the training batches based on the distribution of classes in the training set. These weights differ from our reweighting implementation as it is an offline process with pre-calculated probabilities selected before training while our implementation adjusts the weights during the training process. The comparison between these three approaches will help demonstrate the effectiveness of the learning to reweight parameters algorithm.

The reweighting algorithm was implemented using a similar approach as was previously described in the toy problem in \autoref{sec:ToyImplementation}. Additionally, we extracted two different validation sets for training purposes. The first validation set was a small balanced subset used to train our example weighting, consisting of 3 samples per class. The second validation subset was used to test the generalization of our model after each epoch. This subset was also balanced in order to impartially generalize the overall accuracy of the model with 15 samples per class.

For testing, we utilized the 2018 ISIC challenge's test data set and ground truth. We used a balanced multi-class accuracy function to match the metric used by the challenge's official submission website. This involves averaging the individual accuracy of each class. Our Implementation can be found at the github repository cited in this paper ~\cite{github_repo}.

\subsection{Real-world Problem: Results}
The balanced multi-class accuracy of the different training methods can be seen in Table ~\ref{tab:learning_accuracy}. The performance of our implemented algorithm exceeds the performance of weighted random sampler and the unweighted training by 48.87\% and 51.49\% respectively. The implemented method can be seen to be an effective approach for handling class imbalance. However, we were unable to produce results competitive with the top 10 in the leader-board of the ISIC Challenge in which the cutoff was 78\%, however with more resources, training time, and experimentation with different models the performance may have been able to improve.
\begin{table}[H]
\centering
\caption{Balanced Multi-class Accuracy}
\label{tab:learning_accuracy}
\begin{tabular}{lccc}
\toprule
Model & Learning Method & Accuracy(\%) \\
\midrule
ResNet50 & Learning to Reweight    & 65.56\\
ResNet50 & Weighted Sampler   & 16.69\\
ResNet50 & Unweighted Training   & 14.07\\
\bottomrule
\end{tabular}%
\end{table}

In Table 2 we can see that the learning to reweight algorithm has a better distribution of class accuracy than the other methodologies. The other two training techniques had several classes with a 0\% accuracy, which is most likely attributed to the imbalanced nature of the problem. Our algorithms dynamic adjustment of weights during training seems to be an effective way of dealing with this difficulty. Additionally, our algorithms success can also be attributed to the small balanced validation set that the example weights are trained on.

In figure 3 and 4 we can see the confusion matrices of the unweighted learning algorithm and the learning to reweight algorithm. In these figures we can see the diagonal (correct predictions) of the reweight algorithm is more dominant than in the unweighted algorithm. We can also see that the unweighted algorithm commonly mistook the nv class for bkl and bcc classes while the reweighting algorithm seemed to correct this misconception.
\begin{table}[H]
\centering
\caption{Summary of Testing Accuracy for Different Classes}
\label{tab:class_accuracies}
\resizebox{\columnwidth}{!}{%
\begin{tabular}{lccc}
\toprule
Class & Learning to Reweight (\%) & Unweighted Training (\%) & Weighted Sampler (\%) \\
\midrule
nv    & 74.48 & 7.04 & 86.25 \\
mel   & 69.59 & 0.0 & 0.0 \\
bkl   & 54.38 & 39.63 & 26.27 \\
bcc   & 80.65 & 49.46 & 4.30 \\
akiec & 55.81 & 2.33 & 0.0 \\
vasc  & 62.86 & 0.0 & 0.0 \\
df    & 61.3 & 0.0 & 0.0 \\
\bottomrule
\end{tabular}%
}
\end{table}
\begin{figure}[H]
    \centering
    \includegraphics[scale=0.25]{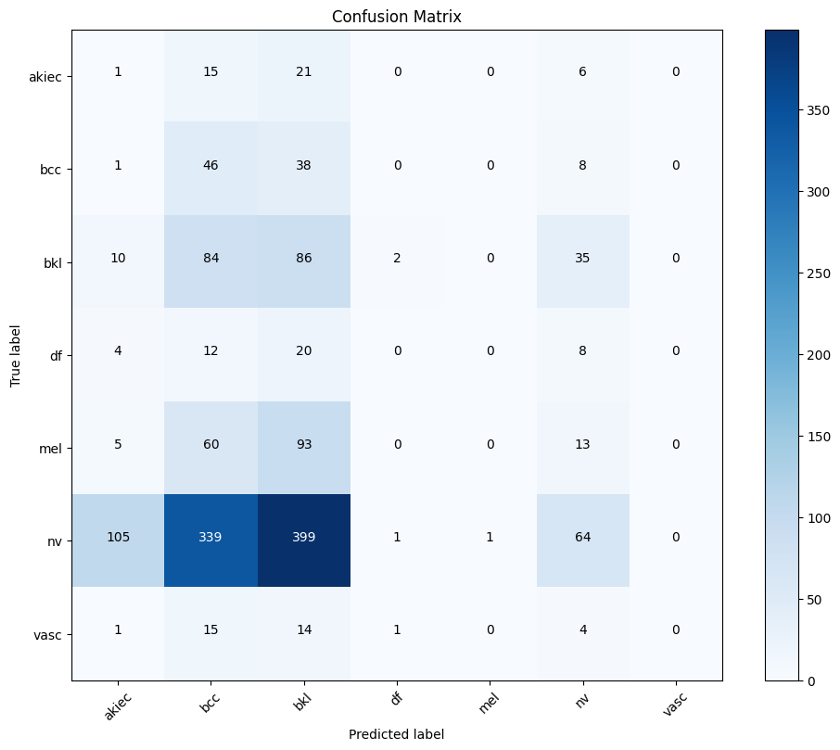}
    \caption{Confusion Matrix for Unweighted Training}
    \label{fig:unweighted_conf}
\end{figure}
\begin{figure}[H]
    \centering
    \includegraphics[scale=0.25]{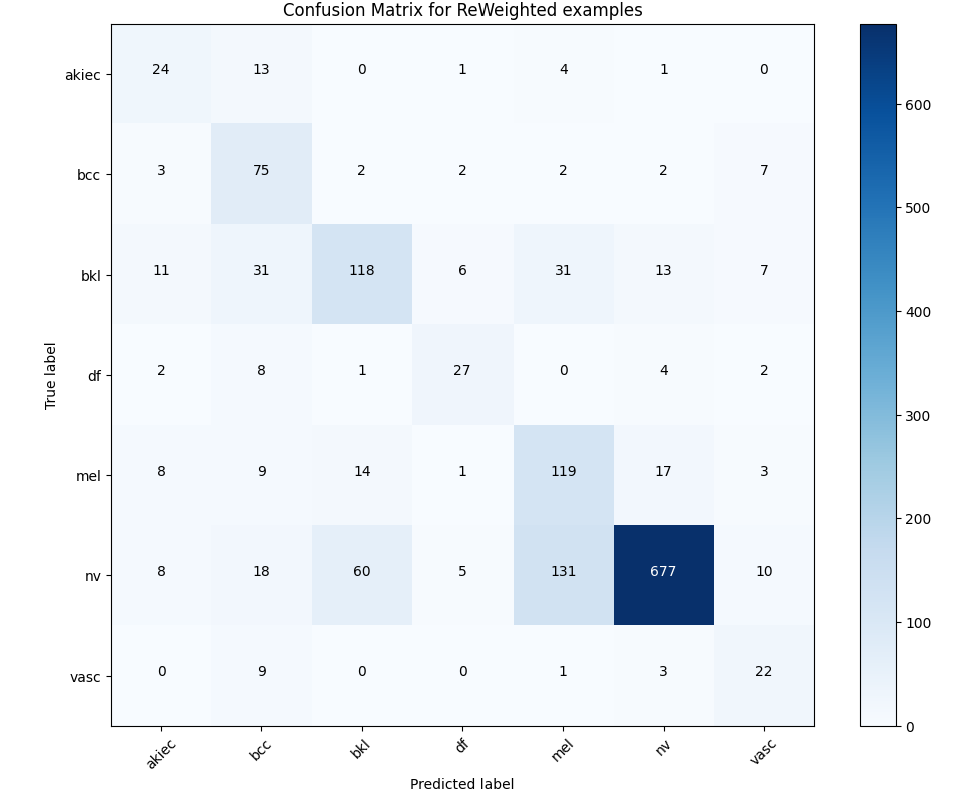}
    \caption{Confusion Matrix for Learning to Reweight Training}
    \label{fig:learning_conf}
\end{figure}

\newpage
\section{Conclusion}
Our results show that the re-weighting algorithm we have re-implemented is able to effectively handle class imbalance. We have demonstrated that this approach significantly outperforms traditional training methods like SGD and Weighted Random Sampling. Potential progressions of our implementation would be to test the effectiveness of handling a dataset with a combination of both label noise and class imbalance and compare it more thoroughly with other possible solutions. The main feature of the real-world problem addressed in this paper is class imbalance, although label noise could have also been introduced into the dataset from incorrect diagnosis caused by human error. Finding a dataset with natural label noise or creating one through random poisoning could help us better see the effects of the reweighting algorithm.

Additionally, our algorithm, as mentioned previously, was unable to achieve competitive results compared to the top performers of the official 2018 ISIC Challenge. This displays the fact that even with a training algorithm that is able to do well against issues like class imbalance and label noise, there are still many other variables and approaches that need to be refined when creating a model. Still, with more resources and experimentation, we believe that our approach could achieve higher accuracy and improve its standing in the leader-board.

In conclusion, our study contributes to the growing body of literature on addressing class imbalance in machine learning applications, particularly in medical diagnosis. By re-implementing and validating the learning to reweight algorithm's findings, we provide a valuable tool for researchers to improve the accuracy and reliability of automated diagnostic systems.

\newpage
\bibliography{bib}

\begin{thebibliography}{9}
\providecommand{\natexlab}[1]{#1}
\providecommand{\url}[1]{\texttt{#1}}
\expandafter\ifx\csname urlstyle\endcsname\relax
  \providecommand{\doi}[1]{doi: #1}\else
  \providecommand{\doi}{doi: \begingroup \urlstyle{rm}\Url}\fi

\bibitem[Com()]{Competition}
{ISIC Challenge 2018}.
\newblock \url{https://challenge.isic-archive.com/data/#2018}.
\newblock Accessed: 2024-04-09.

\bibitem[Finn et~al.(2017)Finn, Abbeel, and Levine]{DBLP:journals/corr/FinnAL17}
Finn, C., Abbeel, P., and Levine, S.
\newblock Model-agnostic meta-learning for fast adaptation of deep networks.
\newblock \emph{CoRR}, abs/1703.03400, 2017.
\newblock URL \url{http://arxiv.org/abs/1703.03400}.

\bibitem[Frénay \& Verleysen(2014)Frénay and Verleysen]{Frenay2014}
Frénay, B. and Verleysen, M.
\newblock Classification in the presence of label noise: A survey.
\newblock \emph{Neural Networks and Learning Systems, IEEE Transactions on}, 25:\penalty0 845--869, 05 2014.
\newblock \doi{10.1109/TNNLS.2013.2292894}.

\bibitem[Gardner et~al.(2024)Gardner, Boardley, et~al.]{github_repo}
Gardner, J., Boardley, B., et~al.
\newblock Ece 50024 final project code.
\newblock \url{https://github.com/jgardner21/MLProject}, 2024.

\bibitem[Kennedy et~al.(2021)Kennedy, Johnson, and Khoshgoftaar]{9643276}
Kennedy, R. K.~L., Johnson, J.~M., and Khoshgoftaar, T.~M.
\newblock The effects of class label noise on highly-imbalanced big data.
\newblock In \emph{2021 IEEE 33rd International Conference on Tools with Artificial Intelligence (ICTAI)}, pp.\  1427--1433, 2021.
\newblock \doi{10.1109/ICTAI52525.2021.00227}.

\bibitem[Malisiewicz(2011)]{MalisiewiczTomasz2011ERfO}
Malisiewicz, T.
\newblock Exemplar-based representations for object detection, association and beyond, 2011.

\bibitem[Ren et~al.(2018)Ren, Zeng, Yang, and Urtasun]{pmlr-v80-ren18a}
Ren, M., Zeng, W., Yang, B., and Urtasun, R.
\newblock Learning to reweight examples for robust deep learning.
\newblock In Dy, J. and Krause, A. (eds.), \emph{Proceedings of the 35th International Conference on Machine Learning}, volume~80 of \emph{Proceedings of Machine Learning Research}, pp.\  4334--4343. PMLR, 10--15 Jul 2018.
\newblock URL \url{https://proceedings.mlr.press/v80/ren18a.html}.

\bibitem[Team19(2024)]{team19}
Team19.
\newblock {Reweighting for Deep Learning}.
\newblock \url{https://github.com/Parth1811/reweighting_for_deep_learning}, 2024.

\bibitem[Tschandl(2018)]{Ham10000_2018}
Tschandl, P.
\newblock {The HAM10000 dataset, a large collection of multi-source dermatoscopic images of common pigmented skin lesions}, 2018.
\newblock URL \url{https://doi.org/10.7910/DVN/DBW86T}.

\end{thebibliography}
\bibliographystyle{icml2023}

\newpage
\section{Appendix}
    \subsection{Theorem 1}
    Assuming the loss function is Lipschitz-smooth with constant L, have $\sigma$-bounded gradients and learning rate $ \alpha_t = \frac{2n}{L\sigma^2} $, it can be shown that,

    \begin{equation}
        G(\theta_t + 1) \leq G(\theta_t)
    \end{equation}

    \begin{equation}
        Where, G(\theta_t) = \frac{1}{M} \sum_{i=1}^M {J^v_i\left( \theta_{t+1}(\epsilon) \right)}
    \end{equation}

    Further, the theorem also states a strong relation for the expectation of the loss function. It says that for any time step, the expectation of loss function is equal to the value of the previous time step if and only if the gradient of loss is equal to zero; that is, convergence only happens at a minima. Mathematically $ E[G(\theta_{t+1}) = G(\theta_t)] \Leftrightarrow \nabla G(\theta_t)=0 $

    The proof for this theorem is two parts. First, the authors use the Lipschitz-smoothness of the loss function to establish this inequality

    \begin{equation}
        G(\theta_t + 1) \leq G(\theta_t) + \nabla G^T \Delta \theta + \frac{L}{2}||\theta||^2
    \end{equation}

    This equation, upon simplifications using the algorithm's $ \Delta \theta $, yields the equation (\ref{th1:in}) below. Here $\tau_t$ is positive and so is $ 1 - \frac{L\alpha_t\sigma^2}{2n} $ because of the assumption regarding learning rate.

    \begin{equation} \label{th1:in}
        G(\theta_t + 1) \leq G(\theta_t) - \frac{\alpha_t}{n}\tau_t\left( 1 - \frac{L\alpha_t\sigma^2}{2n}\right)
    \end{equation}

    Next the authors, use of $\tau_t$ to show that  $ E[\tau_t] = 0 \Leftrightarrow \nabla G(\theta_t)=0 $. This fact is then used to prove the second part of the theorem.
    
    \subsection{Theorem 2}

The following proof is used to determine the convergence rate of the method. If we assume $G$ is Lipschitz-smooth with constant $L$ and the training loss function $f_i$ has $\sigma$-bounded gradients, it can be proven that $\mathbb{E}[\|\nabla G(\theta_t)\|^2] \leq \epsilon$ within $O(1/\epsilon^2)$ steps.

Beginning from the gradient descent update rule:
\begin{equation}
    \theta_{t+1} = \theta_t - \alpha_t \nabla G(\theta_t),
\end{equation}
and $\alpha_t \leq \frac{2n}{L\sigma^2}$ as per the given conditions.

Since $G$ is known to be Lipschitz-smooth,
\begin{equation}
    G(\theta_{t+1}) \leq G(\theta_t) + \nabla G(\theta_t)^\top(\theta_{t+1} - \theta_t) + \frac{L}{2}\|\theta_{t+1} - \theta_t\|^2.
\end{equation}

We can substitute the update rule into the inequality:
\begin{equation}
    G(\theta_{t+1}) \leq G(\theta_t) - \alpha_t \|\nabla G(\theta_t)\|^2 + \frac{L\alpha_t^2}{2}\|\nabla G(\theta_t)\|^2.
\end{equation}

Next we use the bound on $\alpha_t$ to solve for $\|\nabla G(\theta_t)\|^2$
\begin{equation}
    \|\nabla G(\theta_t)\|^2 \leq \frac{2}{\alpha_t(2 - L\alpha_t)}(G(\theta_t) - G(\theta_{t+1})).
\end{equation}

From there, we can take the expectation of both sides. After summing both sides of the equation over $T$ steps and dividing by $T$, we are left with the following result:
\begin{equation}
    \min_{0 < t < T} \mathbb{E}[\|\nabla G(\theta_t)\|^2] \leq \frac{1}{T}\sum_{t=0}^{T-1} \mathbb{E}[\|\nabla G(\theta_t)\|^2],
\end{equation}
Since the right side of the equation is independent of convergence, we can express it as a constant $C$, divided by $\sqrt{T}$. The $\sqrt{T}$ represents the diminishing impact of each step on reducing the gradient norm as T increases.
\begin{equation}
    \min_{0 < t < T} \mathbb{E}[\|\nabla G(\theta_t)\|^2] \leq \frac{C}{\sqrt{T}} .
\end{equation}
Thus, it has been shown that the algorithm can converge to a constant $\epsilon$ in $O(1/\epsilon^2)$ steps.
\end{document}